\documentclass[runningheads]{llncs}
\usepackage{graphicx}
\usepackage{booktabs}
\usepackage{caption}
\usepackage{subcaption}
\usepackage{threeparttable}
\usepackage{caption}
\usepackage{fancyhdr}
\usepackage{url}

\fancypagestyle{TitlePage}{
\lfoot{\emph{Proceedings of the 13th International Conference on Social Robotics (ICSR 2021). Copyright 2021 by the author(s).}}
}

\begin{document}
\title{Migratable AI : Investigating users' affect on Identity and Information migration of a Conversational AI agent}
%
%
\author{Ravi Tejwani\inst{1}, Boris Katz\inst{1} \and
Cynthia Breazeal\inst{1}}
\institute{Massachusetts Institute of Technology, Cambridge, USA
\email{\{tejwanir,boris,breazeal\}@mit.edu}
}
\maketitle              
\thispagestyle{TitlePage} 

\begin{abstract}
Conversational AI agents are becoming ubiquitous and provide assistance to us in our everyday activities. In recent years, researchers have explored the migration of these agents across different embodiments in order to maintain the continuity of the task and improve user experience. 
In this paper, we investigate user's affective responses in different configurations of the migration parameters. We present a 2x2 between-subjects study in a task-based scenario using information migration and identity migration as parameters. We outline the affect processing pipeline from the video footage collected during the study and report user's responses in each condition. Our results show that users reported highest joy and were most surprised when both the information and identity was migrated; and reported most anger when the information was migrated without the identity of their agent. 

\end{abstract}

\section{Introduction}
We are surrounded by conversational AI agents, such as Alexa \cite{alexa}, Jibo \cite{jibo} or Google Home \cite{googlehome}, as they assist us in our daily activities like providing weather and news updates, ordering meal and ride shares, setting room temperature  etc. These agents build model of our personal preferences and interests as we interact and develop relationship with them. We also interact with the robotic agents in public setting, such as Pepper \cite{pepper}, Kuri \cite{kuri}, or Moxi \cite{moxi} at hospitals, restaurants, and grocery stores, where we share our preferences with them. 
Since these agents exist in different form factors or embodiments and setting, they do not always share information amongst each other. However, the migration of information or identity across embodiments could lead to changes in users' perception \cite{tejwani2020migratable} and affective states. For example, after your interaction with the home agent, Alexa, you might be surprised to see if the restaurant robot greets you with your name and knows about your food order when you enter the restaurant.

Agent migration is a concept which allows an agent to disembody from its current form and migrate to different embodiments while maintaining the relationship with the user. 
Prior work has explored the concept of agent migration through various different architectures\cite{imai1999agent} \cite{lirec} \cite{aylettbody}  \cite{duffy2003agent}. They explored the migration in the form of a synthetic character or a visual entity than compared to a conversational AI agent. Further, several user studies on agent migration \cite{readymigration} \cite{kriegel2010digital} \cite{grigore2016comparing} explored users impression on the agent such as validating if the users perceived that it was the same agent in another embodiment, or if the users understood the concept of agent migration. For instance, Syrdal et. al. performed series of group discussions with a school class children, aged 3 to 6, on evaluating children's impressions on the understanding of the concept of migration\cite{syrdal2009boy}. However, the affective behavior analysis of the users in the context of migration of AI agents have not been studied before. User's affective behavior and autonoumous reactions provide a deeper understanding of user's reaction towards the system in comparison to the subjective reports given by the users \cite{roz} \cite{Mello} \cite{Sam} which would be beneficial in designing effective migratable AI agents.

In our previous work \cite{tejwani2020migratable}, we proposed a Migratable AI system which allows a conversational AI agent to migrate across different physical embodiments. We measured the user's perception on trust, competence, likability and social presence using information migration and identity migration as parameters. In this paper, we build upon our previous work by analyzing the affective behavior of the users during the migration of the conversational AI agent. We ran a  2x2 between-subjects study in a task-based scenario using information migration and identity migration as parameters to investigate the affective responses of the users. We outline the affect processing pipeline from the video footage collected during our experimental study. The pipeline comprised of two stages: affect detection and affect interpretation. The findings from this paper, can be used for the further development of effective migratable systems.

\section{Related Work}

\subsection{Identity migration}
Prior work has explored the concept of agent migration through various different identity migration architectures \cite{aylettbody} \cite{lirec} \cite{duffy2003agent}.  The agent migration was first explored by Imai et al. \cite{imai1999agent} by demonstrating a tour guide application where a personal agent could migrate from mobile device to a physical robot. Later, the research by Martin et al. \cite{martin2005maintaining} explored that the identity is not just "Who am I?" but "Who am I in the eyes of others?" where they proposed the visual identity cues in their experiment that characters share a common feature - such as a hat or glasses, common colour scheme, common set of markings, or characters are of the same class of objects.  Further,  \cite{readymigration} \cite{grigore2016comparing} explored the research questions such as ``Do participants feel that they are interacting with the same agent across different embodiments?''; ``What are the most important aspects of an agent to communicate identity retention?''; or ``Do users perceive the agents in different embodiments as the same entity?"

\subsection{Information migration}
Information migration architectures were explored in \cite{lirec} \cite{ono2000reading} as generic memory models and persistent memory models. Aylett et al. proposed CMION in  \cite{aylettbody} \cite{lirec}, an open source architecture comprised of three layers (Mind, Mind-Body, and Body), that served as a framework for the bidirectional mapping of information to different levels of abstraction (i.e., from raw sensory data to symbolic data and vice versa). These models were created to focus on the following three different aspects: 

\begin{enumerate}
\item \textbf{Scope} - Short term memory (STM) was modeled computationally to maintain a companion's current focus and Long term memory (LTM) was used for the artificial companions that interact with human users over a long period of time

\item \textbf{Efficiency} - how to optimize the storage and recall of memory contents; forgetting through the processes of generalization and memory restructuring.

\item \textbf{Adaptability} - how to use different conversational strategies for information or memory that the robot remembers during the interaction with human (no-memory, partial memory, complete memory). 
\end{enumerate}

\subsection{User Perception of Agent Migration}
User studies have explored users' perception of agents that can migrate across forms in \cite{kriegel2010digital} \cite{readymigration} \cite{grigore2016comparing}  by studying the higher level users' impression on the agent such as validating if the users felt that it was the same agent in another embodiment, or if the users understood the concept of agent migration. Further, \cite{kriegel} \cite{duffy2003agent} explored the users perception on the long term interaction derived from the companion's interaction history both with the environment and the user. 

In this paper, we go beyond the users perception on agent's identity or the subjective reports and investigate user's affective state during the migration of an conversational AI agent using information migration and identity migration as parameters.

\begin{table}[h]
  \centering
   \caption{Participant Demographics}
  \begin{tabular}{ccccc}
    \hline
    Condition&Female&Male&Other&Age(Std. Dev.)\\
    \hline
    (INF+,ID+)&8&10&0&24.4(5.06)\\
    (INF+,ID-)&9&9&0&24.6(6.09)\\
    (INF-,ID+)&7&10&1&28.2(10.2)\\
    (INF-,ID-)&8&9&1&22.6(3.61)\\
  \hline
\end{tabular}
 \label{tab:participant-demographics}
\end{table}

\section{Method}
\subsection{Participants}
We recruited 72 participants from Cambridge area using email advertisements. Participants were between 18 and 54 years old with mean age M=24.2, SD=5.09. Participants were randomly assigned and counterbalanced by gender across the four conditions (n=18 per condition) as described in Table~\ref{tab:participant-demographics}. The study was approved by our Institutional Review Board, and participants signed an informed consent form prior to the study. 

\subsection{Study Protocol}
\begin{figure}[h]
    \centering
    \scalebox{0.5}{
        \includegraphics[width=\linewidth]{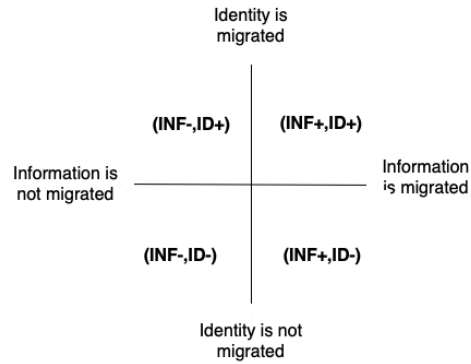}
    }
    \caption{Study conditions}
    \label{fig:study-conditions}
    \vspace{-6mm}
\end{figure}

We ran a $2 \times 2$ between-subjects study with \emph{\textbf{Information migration $\times$ Identity migration}}. The 4 conditions used in the study are described in Figure ~\ref{fig:study-conditions}.

We used the Migratable AI system~\cite{tejwani2020migratable} in the study. The system allowed the conversational AI agent to migrate across different embodiments by preserving its identity(identity migration) and/or remembering the information context (information migration). Each participant began the study in our lab's study room, modeled as their "home", with the home agent (Alexa)~\cite{alexa}. The home agent delivered the participant's schedule for the day which included a job interview. Throughout the conversation, the home agent learned about the participant such as his/her name and feelings.

The mobile robot (Kuri)~\cite{kuri} was located in a hallway of the lab which played the role of front desk receptionist robot at the interview location. The receptionist robot, changed its appearance to look and sound like home agent (when identity was migrated) or continued to look and sound like Kuri with a different voice profile (when identity was not migrated). The receptionist robot detected their face, recognized the participant by name, and acknowledged the reason for their visit (when information was migrated) or prompted the participant for their name and reason for their visit (when information was not migrated). During the conversation, the receptionist robot either validated the participant's feelings (when information was migrated) or asked how they were feeling for their interview (when information was not migrated). The receptionist robot also learned the participant's drink preferences (coffee, water or tea) and escorted the participant to the interview waiting area.

\begin{figure}[t]
   \centering
   \captionsetup{justification=centering}
    \scalebox{0.50}{
    \includegraphics[width=\linewidth]{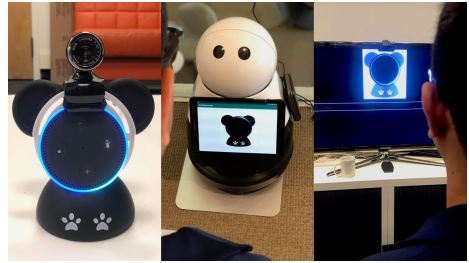}
    }
    \caption{Left to right: Home agent, Home agent migrating to receptionist robot, Home agent migrating to waiting room assistant.}
    \label{fig:teaser}
     \vspace{-6mm}
\end{figure}

At the interview waiting area, the participant interacted with the waiting room assistant (Smart TV) which conversed with the participant until the arrival of the interviewer. It changed its appearance to look and sound like home agent (when the identity was migrated) or continued to look and sound like itself (when the identity was not migrated). While the participant waited, it offered the participant their preferred drink (which it remembered in the condition when the information was migrated) or offered the participant a drink while waiting (when the information was not migrated). It also acknowledged the participant's feelings (when the information was migrated) and wished them good luck before the interviewer arrived. The interviewer role was enacted by the experimenter.

For identity migration - the design decisions were informed from the past literature on what helps users perceive an identity of an agent \cite{aylettbody} \cite{martin2005maintaining} \cite{cuba2010agent}. In the identity migration conditions, the same visual characteristics (Figure ~\ref{fig:teaser}, panda-esque circular appearance) and voice (Joanna TTS) was used across all embodiments to convey identity continuity.

For information migration -  the information parameters such as the person's name, feelings about the interview, drink preference and reason for visit were learned by each agent during the conversation. If the system was configured to migrate information across embodiments, this information was shared amongst the agents to maintain the continuity of the interaction else the agent had to prompt the user for the information. The number of conversational turns (four in this user study) touch basing the personal and non-personal information between the agent and the participant were kept consistent across all the conditions. This might necessitate the agent to repeat certain questions to the users when the information was not migrated but it was to ensure that we do not create a bias in the study and keep the conversational turns consistent. 


\begin{figure}%
    \centering
    \subfloat[\centering Raw affect data of a participant]{{\includegraphics[width=5cm]{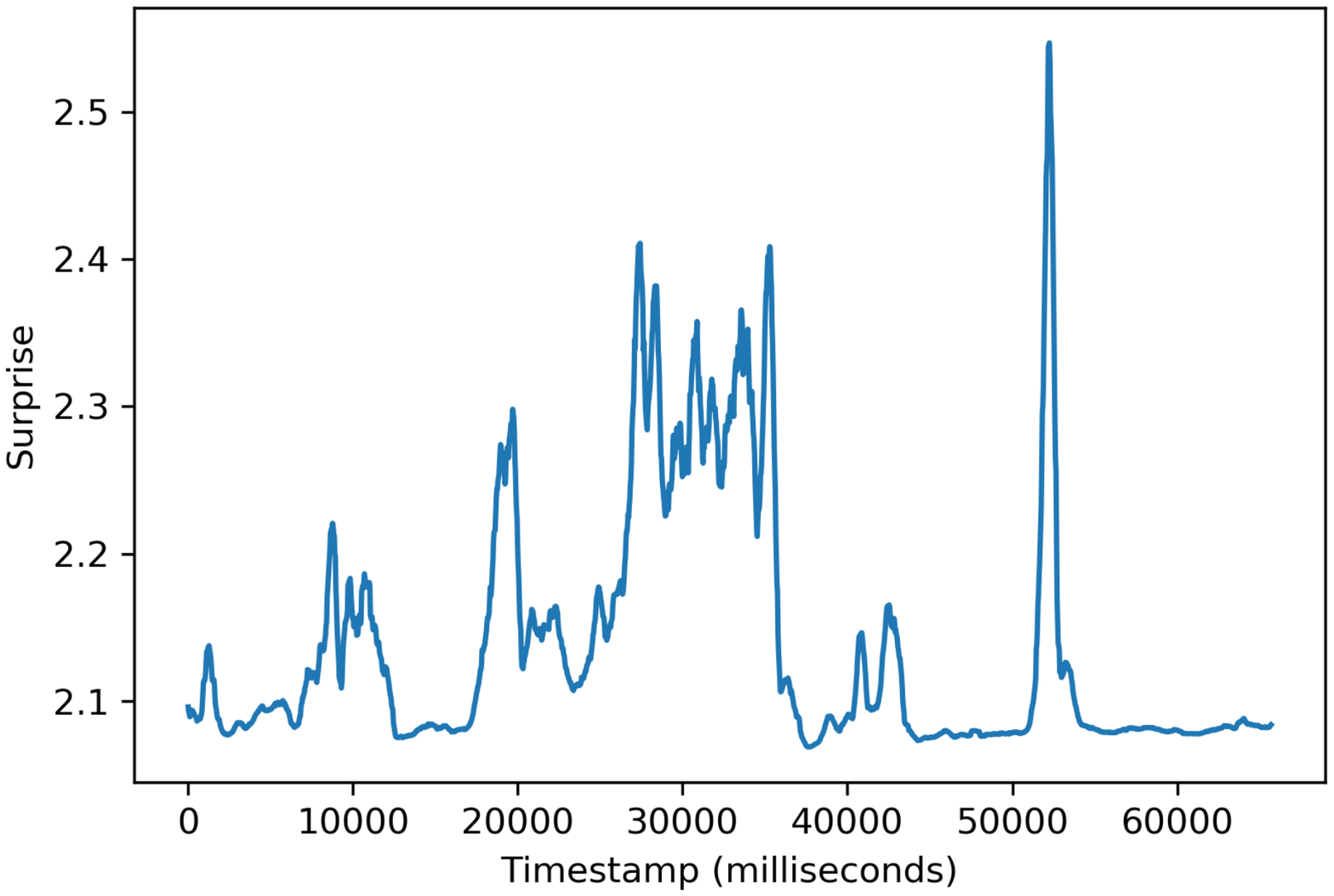} }}%
    \qquad
    \subfloat[\centering Affect data after applying median filter and threshold]{{\includegraphics[width=5cm]{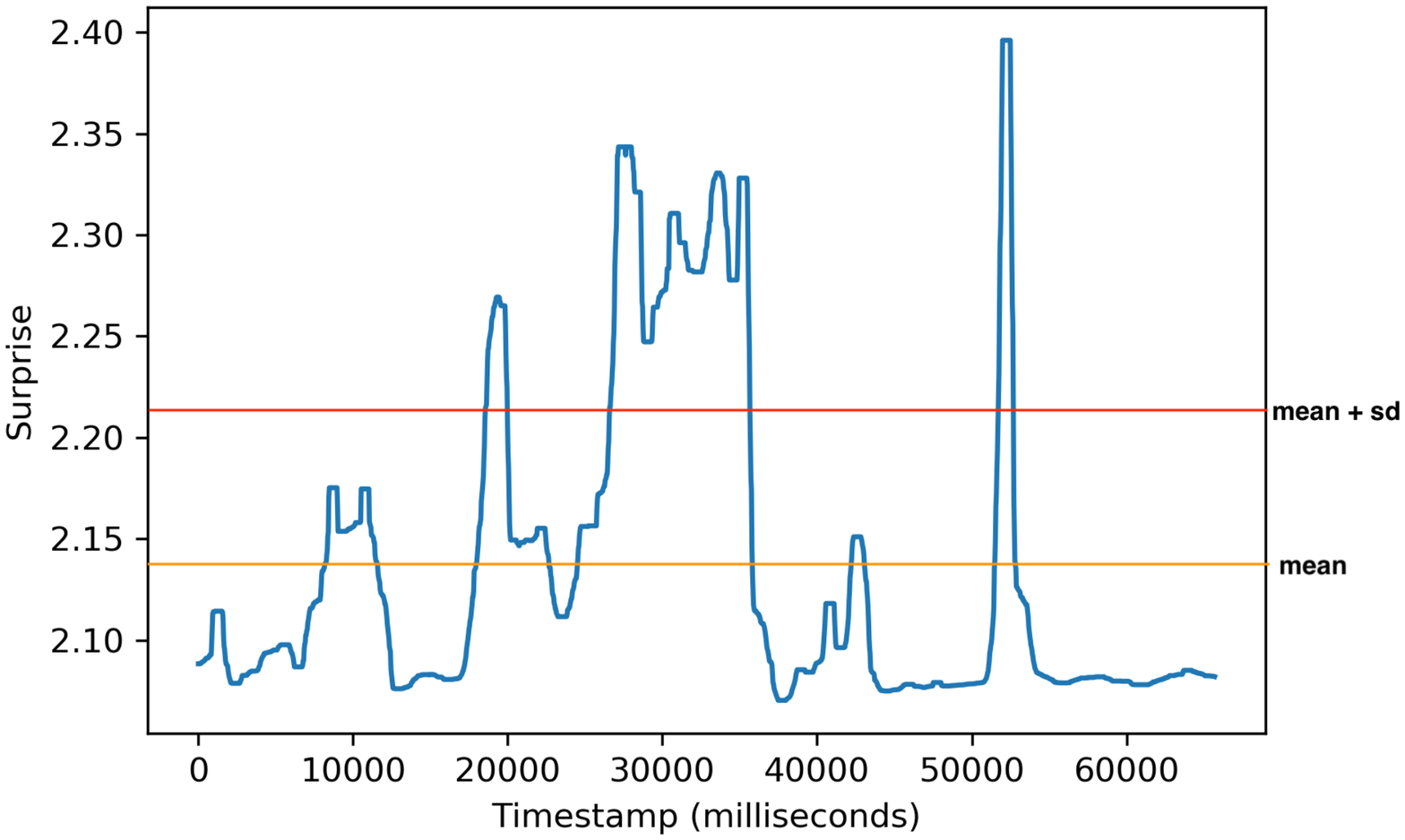} }}%
    \caption{Affect interpretation for surprise of a sample participant}%
    \label{fig:affect-interpretation}%
    \vspace{-4mm}
\end{figure}

\subsection{Data Collection}

A front facing USB camera were connected to a Raspberry Pi and attached to each embodiment to record the interaction. The Raspberry Pi used face detection to send a wake up signal to the robot/embodiment. It began recording the video when the participant's face was detected in front of the embodiment and stopped the recording when the interaction with the user ended. The video recordings of all the 72 participants (18 in each condition) were processed using Affdex~\cite{affdex} for affect analysis.

\begin{table*}[]
\small
\centering
\caption{Affect features by each migration condition \newline}
\label{tab:my-table}
\resizebox{\textwidth}{!}{%
\begin{tabular}{|c|c|c|c|c|}
\hline
\textbf{Features} & \textbf{(INF+,ID+)} & \textbf{(INF+,ID-)} & \textbf{(INF-,ID+)} & \textbf{(INF-,ID-)} \\ \hline
\textbf{joy}               & $0.274\pm0.03$ & $0.188\pm0.01$ & $0.256\pm0.02$ & $0.239\pm0.02$ \\ \hline
\textbf{anger}             & $0.179\pm0.01$ & $0.216\pm0.02$ & $0.151\pm0.01$ & $0.157\pm0.02$ \\ \hline
\textbf{surprise}          & $0.181\pm0.01$ & $0.144\pm0.02$ & $0.139\pm0.02$ & $0.157\pm0.02$ \\ \hline
\textbf{smile}             & $0.254\pm0.03$ & $0.274\pm0.04$ & $0.306\pm0.03$ & $0.187\pm0.02$ \\ \hline
\textbf{brow\_raise}       & $0.226\pm0.02$ & $0.188\pm0.01$ & $0.261\pm0.02$ & $0.231\pm0.02$ \\ \hline
\textbf{brow\_furrow}      & $0.201\pm0.02$ & $0.201\pm0.01$ & $0.252\pm0.03$ & $0.269\pm0.04$ \\ \hline
\textbf{nose\_wrinkle}     & $0.258\pm0.03$ & $0.222\pm0.02$ & $0.283\pm0.03$ & $0.298\pm0.04$ \\ \hline
\textbf{upper\_lip\_raise} & $0.230\pm0.02$  & $0.219\pm0.02$ & $0.219\pm0.01$ & $0.215\pm0.02$ \\ \hline
\textbf{mouth\_open}       & $0.226\pm0.01$ & $0.205\pm0.02$ & $0.257\pm0.03$ & $0.229\pm0.02$ \\ \hline
\textbf{eye\_closure}      & $0.214\pm0.01$ & $0.25\pm0.02$  & $0.253\pm0.02$ & $0.289\pm0.03$ \\ \hline
\textbf{cheek\_raise}      & $0.204\pm0.04$ & $0.162\pm0.02$ & $0.315\pm0.03$ & $0.278\pm0.04$ \\ \hline
\end{tabular}%
}
\caption*{
\textbf{ID+ or ID-} represents identity is migrated or not migrated. \newline
\textbf{INF+ or INF-} represents information is migrated or not migrated.}
\end{table*}

\section{Affect processing pipeline }

Affect analysis has been performed in the past using several  statistical heuristics such as mean value of the pertaining window ~\cite{Jeong}, if at any given point in the window the value of the metric exceeds a given threshold ~\cite{Zhang}, or if the mean value of the metric over pertaining window exceeds a given threshold ~\cite{Bernin}. We implemented the pipeline for affect detection and interpretation using smoothing and a threshold technique. Most of our pipeline overlaps with the approach detailed by Spaulding and Breazeal~\cite{Sam} and D’Mello, Kappas, and Gratch~\cite{Mello}.

The data pipeline is as follows: Raw data, \textit{RD = $rd_{0}$, $rd_{1}$, ...., $rd_{n}$}, is used for an interaction time window W, where \textit{$|RD| >> W$} and $rd_{x}$ is $x$ participant's raw data. The raw data, RD, is further processed by an affect detector which produces feature vectors of metrics, \textit{M = $m_{0}$, $m_{1}$, ...., $m_{n}$} (e.g., the degree to which ‘joy’ ‘smile’, ‘brow raise’ etc. are expressed), for each data point. The affect interpreter further analyzes these metric vectors for the time window and produces a feature label, l, for that window. 

\subsection{Affect Detection}
The facial expressions of the participants were evaluated from the video captured by the front facing USB camera mounted on each of the embodiments at 30fps. The camera would get activated at the detection of the participant's face and record the video for the time frame of the interaction between the agent and the participant at each embodiment (Figure ~\ref{fig:teaser}). We processed the frames from each embodiment, using the Affdex ~\cite{affdex} as the affect detector, which detected features such as: \textit{'joy', 'anger', 'surprise', 'smile', 'brow\_raise', 'brow\_furrow', 'nose\_wrinkle', 'upper\_lip\_raise', 'mouth\_open', 'eye\_closure', 'cheek\_raise'}.

\subsection{Affect Interpretation}
The affect data for the interaction duration between the agent and participant on an embodiment, is collected and converted to a feature vector. We interpret each feature as a binary indicator variable whose value is determined by smoothing and a threshold technique. The raw affect data, $M$, is initially passed through a median-filter smoothing. Further, if the maximum value of the median-smoothed detected peaks exceeds the threshold, then the feature value is interpreted with an indicator value of 1. For the given time window, the interpreted affect feature vector is comprised of set of observed feature indicators. The threshold value for the feature is set at the mean value of the feature across the entire time window plus a standard deviation (Figure ~\ref{fig:affect-interpretation}). Finally, each of the affect feature score is further normalized for the data analysis.

\begin{figure}[t]
  \centering
  \captionsetup{justification=centering}
    \scalebox{0.70}{
    \includegraphics[width=\linewidth]{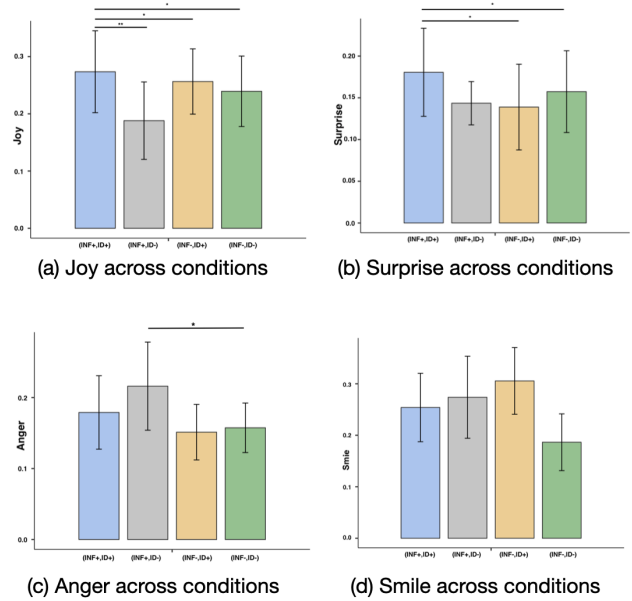}
    }
    \caption{Box-plot for the normalized affective measures across conditions. The boundary of the box closest to zero indicates the 25th percentile, the line within the box marks the mean and the boundary of the box farthest from zero indicates the 75th percentile. Whiskers above and below the box indicate the 10th and 90th percentiles. * means p$<$.05, ** means p$<$.01}
    \label{fig:measures-across-conditions}
    \vspace{-4mm}
\end{figure}    
    
\section{Results}
Normality was first checked for the affective measures from the visual inspection of Q-Q plots and Shapiro-Wilk's test. With all p-values $<$ 0.05, the Shapiro-Wilk test rejects the null hypothesis of data normality, hence, we perform the Kruskal-Wallis H test over the data. Furthermore, the Dwass-Steel-Critchlow-Fligner test was used for the pair-wise comparisons. All the statistical analysis were performed using R(version 3.6.1) and Jamovi\cite{jamovi}. 

We found a significant effect in joy, surprise and anger amongst the users when configuration of the identity and information of the agent was changed during the migration. 

\textbf{Joy:} There was a statistically significant difference in joy scores across different conditions, ${\chi}^2(3)$ = 7.560, p = 0.016. The pair wise comparisons showed that mean joy score of $0.274\pm0.03$ for (INF+, ID+) was significantly greater than $0.188\pm0.01$ for (INF+,ID-) with p=0.008, $0.256\pm0.02$ for (INF-,ID+) with p=0.024 and $0.239\pm0.02$ for (INF-,ID-) with p=0.039 (Figure ~\ref{fig:joy}).

\textbf{Surprise:} There was a statistically significant difference in surprise scores across different conditions, ${\chi}^2(3)$ = 4.40, p = 0.033. The pair wise comparisons showed that mean surprise score of $0.181\pm0.01$ for (INF+, ID+) was significantly greater than $0.139\pm0.02$ for (INF-,ID+) with p=0.036 and $0.157\pm0.02$ for (INF-,ID-) with p=0.042 (Figure ~\ref{fig:surprise}).

\textbf{Anger:} There was a statistically significant difference in anger scores across different conditions, ${\chi}^2(3)$ = 3.157, p = 0.041. The  pair wise comparisons showed that mean anger score of $0.216\pm0.02$ for (INF+,ID-) was significantly greater than $0.157\pm0.02$ for (INF-,ID-) with p=0.032 (Figure ~\ref{fig:anger}).

\textbf{Other affective measures:} The analysis results for the other affective scores were not significantly different across the different conditions: smile (${\chi}^2(3)$ = 2.986, p = 0.091), brow\_raise (${\chi}^2(3)$ = 2.602, p = 0.080), brow\_furrow (${\chi}^2(3)$ = 1.391 p = 0.842), nose\_wrinkle  (${\chi}^2(3)$ = 1.462, p = 0.924), upper\_lip\_raise  (${\chi}^2(3)$ = 2.753, p = 0.178), mouth\_open  (${\chi}^2(3)$ = 1.115, p = 0.273), eye\_closure (${\chi}^2(3)$ = 1.394, p = 0.307) and cheek\_raise (${\chi}^2(3)$ = 2.171, p = 0.092). 

Table ~\ref{tab:my-table} summarizes the mean scores with their standard deviation for all the affective features across all the conditions.

\section{Discussion and Conclusions}
We presented the results from one of the first systematic investigations of users affective behavior on the migration of the conversational AI agent. We ran a 2x2 between-subjects study in a task-based scenario with 72 participants using information migration and identity migration as parameters to investigate the affective behavior of the users. We outlined an affect processing pipeline from the video footage collected during the study. 

We inferred that users expressed most joyfulness and surprise when they saw their agent in a different embodiment and the agent remembered their preferences and context \textbf{(both the information and identity of the agent was migrated)}. This was corroborated by the participant's comments during the post-study interview. Participant P21 said  \textit{``I think it remembers that I am anxious about the interview. That means it cares about me and makes it different from, for example, a coffee machine.''} Another participant, P34, said, \textit{``Being familiar with Alexa, allowed me to trust Receptionist and TV agent''}.

The users were most disappointed and angry when they found out that their information was shared with different agents \textbf{(information was migrated but identity was not migrated)}. P51 said \textit{"I did not trust the agents well because they seemed to share all of the information about me, and I did not want to disclose more."}. Also, P39 said \textit{"... especially after the receptionist agent knew what I told Alexa, I no longer trusted Alexa"}. The insights gained from this research could be used for further development of affective migratable systems. 

\bibliographystyle{splncs04}
\bibliography{main}
\end{document}